\documentclass{article}

\usepackage{algorithmic}
\usepackage{algorithm}
\usepackage{booktabs} 
\usepackage{siunitx} 
\usepackage{multirow} 
\usepackage[table]{xcolor} 
\usepackage{graphicx} 
\usepackage{caption} 
\usepackage{threeparttable} 

\usepackage{amssymb}

\usepackage{amsmath}

\usepackage{color,soul}
\usepackage{tabularx}
\usepackage{multirow}
\usepackage{colortbl}

\usepackage{booktabs} 
\usepackage{siunitx} 
\usepackage{microtype}
\usepackage{graphicx}
\usepackage{subfigure}
\usepackage{booktabs} 

\usepackage{hyperref}

\usepackage[accepted]{mlsys2025}

\mlsystitlerunning{LayerCollapse: Adaptive compression of neural networks}
\newcommand{\MethodName}{LayerCollapse }

\begin{document}

\twocolumn[
\mlsystitle{LayerCollapse: Adaptive compression of neural networks}



\mlsyssetsymbol{equal}{*}

\begin{mlsysauthorlist}
\mlsysauthor{Soheil Zibakhsh Shabgahi}{to}
\mlsysauthor{Mohammad Sohail Shariff}{to}
\mlsysauthor{Farinaz Koushanfar}{to}
\end{mlsysauthorlist}

\mlsysaffiliation{to}{Department of Electrical and Computer Engineering, University of California San Diego, San Diego, USA}
\mlsyscorrespondingauthor{Soheil Zibakhsh Shabgahi}{szibakhsh@ucsd.edu}

\mlsyskeywords{Machine Learning, MLSys}

\vskip 0.3in

\begin{abstract}
Handling the ever-increasing scale of contemporary deep learning and transformer-based models poses a significant challenge. Overparameterized Transformer networks outperform prior art in Natural Language processing and Computer Vision. These models contain hundreds of millions of parameters, demanding significant computational resources and making them prone to overfitting on down stream tasks.
In this work we present \MethodName, a novel structured pruning method to reduce the depth of fully connected layers. We propose an innovative regularizer that promotes shallow fully connected layers, compressing the model with minimal performance impact.
This regularizer enables post-training compression without fine-tuning while preserving performance.
\MethodName controls model expressiveness by regularizing the activation functions between fully connected layers, modulating them to linearity. A linear activation function collapses the rank of a transformation to the rank of the corresponding linear transformation, which demands less resources from the hardware. We demonstrate the effectiveness of \MethodName by showing its compression capabilities in sentimental analysis, text generation, and image classification benchmarks.

\end{abstract}

]



\printAffiliationsAndNotice{\mlsysEqualContribution} 

\section{Introduction}
\label{sec:intro}

The expanding scope of Deep Learning (DL) applications has led to an escalating demand for more accurate models, resulting in an ongoing rise in their complexity. This surge in complexity, however, presents a challenge: It increases the execution costs, in terms of memory and latency. This makes it difficult to deploy these models in real-world scenarios on standard hardware.
Recent studies have shown that modern neural networks often contain unnecessary redundancies, which can be eliminated without affecting their performance. This discovery has spurred interest in model compression techniques such as unstructured pruning \cite{li2016pruning} \cite{jiang2023pruning}, structured pruning \cite{fan2019reducing}, quantization \cite{hubara2017quantized} \cite{jacob2018quantization} \cite{nguyen2020quantization} \cite{krishnamoorthi2018quantizing}, and model distillation \cite{hinton2015distilling}. These methods aim to identify and remove the redundant elements effectively. However, despite significant advances in model compression, implementing the pertinent optimizations typically requires additional steps such as fine-tuning. This can be resource-intensive, making the compression techniques less practical. Moreover some of these methods like unstructured pruning require support from hardware and software to take advantage of the compression \cite{sundaram2023freflex}.

\begin{table}[t]
\centering
\caption{Percentage of model parameters and MACs associated with MLP layers of streamline architectures.}
\label{tab:mlp_contribution}
\begin{tabular}{
  l
  S[table-format=2.1]
  S[table-format=2.1]
}
\toprule
{Model} & {\% Params} & {\% MACs} \\
\midrule
GPT2-Large & 53.4 & 52.7 \\
Bert-Large & 52.6 & 57.1 \\
ViT L/16 & 57.9 & 58.2 \\
Mixer L/16 & 86.5 & 86.5 \\
VGG 16 & 71.2 & 0.6 \\
\bottomrule

\end{tabular}
\end{table}

Multi-Layer Perceptrons (MLPs) \cite{haykin1994neural} are a key architectural component of modern deep neural networks. MLPs are regarded by many as the first contribution to deep learning, present in modern deep learning designs such as the transformer architecture. 
In its simplest form, a $K$-layer is defined as as $K$ linear transformations separated by non-linear operations. MLPs are central to the success of diverse deep learning architectures including encoder only transformer models \cite{vaswani2017attention} \cite{lan2019albert} \cite{raffel2020exploring}, causal transformer models \cite{radford2019language} \cite{touvron2023llama} \cite{gunasekar2023textbooks}, Vision Transformers \cite{dosovitskiy2020image}, and MLP-Mixers \cite{tolstikhin2021mlp}. 
Due to the dense nature of MLPs, a significant portion of a model's computational overhead and storage requirements can be attributed to its MLP layers. 
For instance, MLPs constitute over 60\% of the model's total parameters for the popular architectures discussed in \cite{dosovitskiy2020image} \cite{tolstikhin2021mlp} \cite{simonyan2014very}. 
MLPs contain non-linear activations, e.g., ReLU, which separate  sequences of linear transformations. If the non-linear layers were absent, 
the consecutive linear layers could be condensed into a single linear transformation.
Table \ref{tab:mlp_contribution} showcases the portion of storage and computation requirements of MLP layers is the mentioned architectures.

\begin{figure}[]
  \centering
   \includegraphics[width=1.0\linewidth]{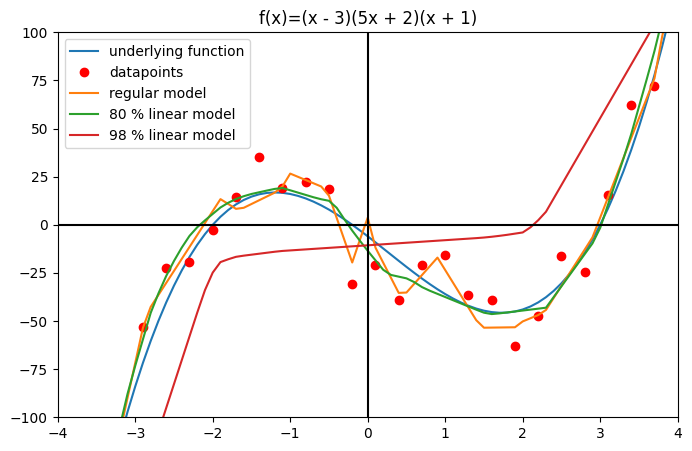}

   \caption{Illustration of regression performance on a two-layer neural network, demonstrating the impact of varying levels of ReLU activation linearity on model overfitting and underfitting. We define the percent of linearity to be the negative slope of the ReLU activation.}
   \label{fig:1d_demo}
\end{figure}

\begin{figure*}[t]
  \centering
   \includegraphics[width=1\linewidth,height=0.5\linewidth, scale=0.8]{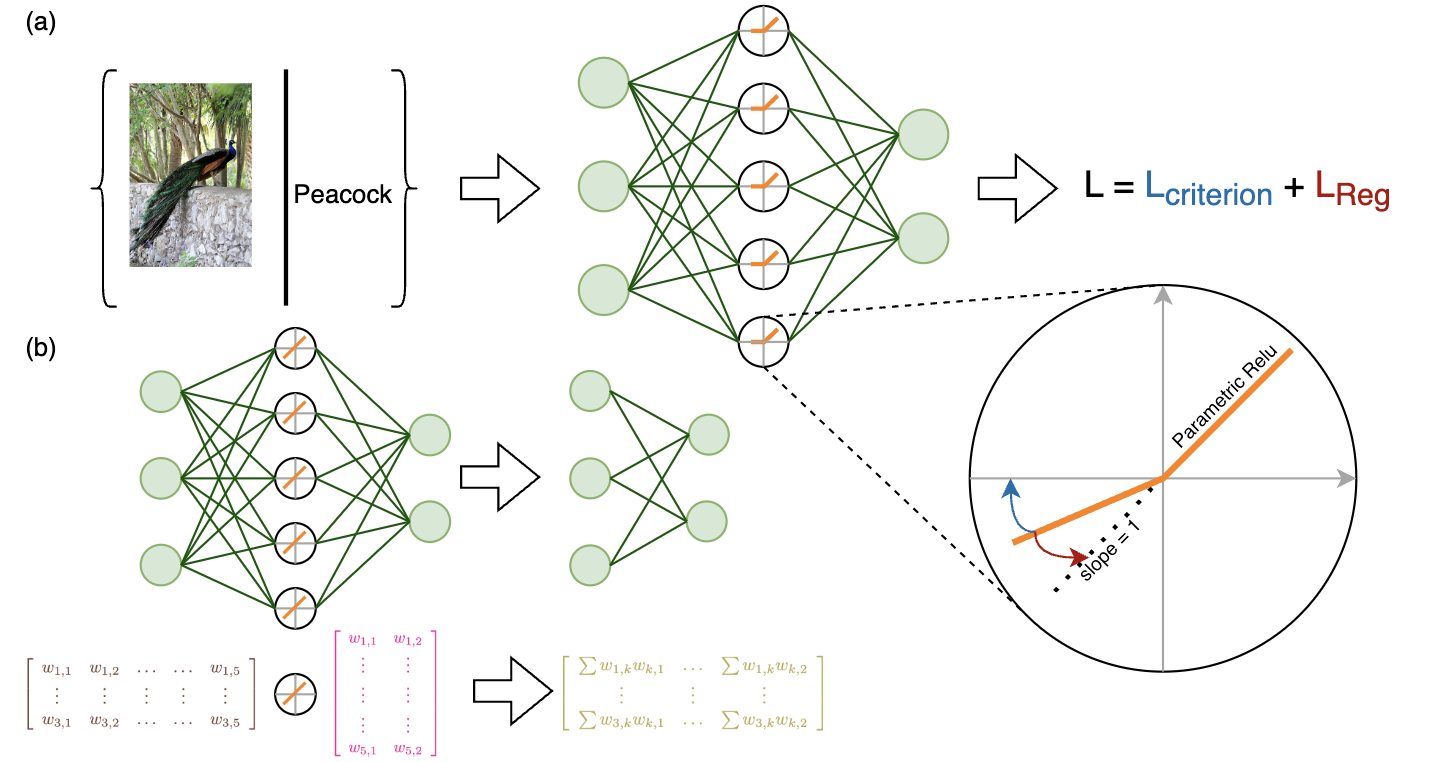}

   \caption{Overview of the LayerCollapse process: $(a)$ The regularization loss shifts the PReLU towards linearity. $(b)$ The layer collapse operation involves eliminating a layer by replacing the hidden layer and adjusting the weights through the product of two matrices.}
   \label{fig:main_idea}
\end{figure*}

In this work, we focus on the non-linear activations inside MLPs. We argue that the expressiveness of an MLP is primarily governed by its non-linear component. By modulating these non-linear characteristics, we can orchestrate a continuum of model expressiveness, ranging from highly complex to near-linear transformations. To gain intuition on the effect of non-linear activations control on model expressiveness, we use a simple 2-layer MLP for a simple regression task. The results are illustrated in figure \ref{fig:1d_demo}. Here an $n\%$ linear model is defined as a model where the negative slope of the leaky ReLU is $n\%$ of one. So, the regular model is just a 2-layer MLP with ReLU activation and a $98\%$ linear model is a 2-layer MLP with a leaky ReLU activation with a negative slope of $0.98$. As we can see, different $n$ values create a continuum of model expressiveness. The regular model overfits to the noise within the data, however, the $80\%$ linear model fits the underlying distribution nicely while the $98\%$ linear model under-fits the data and cannot capture the complexity of the regression task.

A linear transformation can be performed using only a single matrix multiplication, while a $K$-layer MLP requires $K$ matrix multiplications. By adjusting the non-linear component towards linearity, we can regulate the model expressiveness while reducing the operational complexity of the model. We replace two matrix multiplications and a non-linear transform with only one smaller matrix multiplication when the modulated non-linearity is close to linear. In section \ref{sec:methodology}, we will formally define close to linear and its effect on accuracy.

Controlling the model expressiveness using this technique will allow us to tailor the complexity of our model based on the task. Starting from the same architecture, some tasks have small datasets that require a less expressive model to prevent overfitting.

In this work, we developed a novel regularization technique to limit the model expressiveness to prevent overfitting, while reducing the computation cost of the model.


Common, regularization strategies such as Dropout \cite{srivastava2014dropout}, $\ell_1$ and $\ell_2$ norms, have been employed to restrict model expressiveness, attempting robust generalization and preventing overfitting. These regularizers change the model weight structures such that post training compression like pruning and quantization pared with finetuning will yield efficient high performing models.
Fan et al. introduced LayerDrop, a regularization technique to allow for post-training compression without fine-tuning \cite{fan2019reducing}. LayerDrop employs a regularizer that controls the amount of bypass signals in each MLP layer within transformer architectures. Extending this idea, we introduce \MethodName regularization. The \MethodName regularizer works by modulating the negative slope of the ReLU non-linearity to one, effectively transforming it into a linear transformation. This modification not only facilitates the fusion of successive linear layers, conserving computational resources, but also can enhance model generalizability.

This paper presents several key contributions to the field of neural network regularization and compression:
\begin{itemize}
    
    
    
    
    
    \item We introduce \MethodName, a novel regularization approach that enables effective one-shot compression after training.
    \item We make a theoretical analysis providing bounds on the compression loss associated with \MethodName.
    \item We propose a compression-aware regularization loss designed to encourage simpler model structures during the training and finetuning phase.
    \item Through experiments across various datasets in both vision and NLP fields, we demonstrate that \MethodName achieves post-training compression rates of up to 74\% for VGG models and 16\% for transformer architectures. These experiments also highlight enhanced model generalization and reduced overfitting.
    \item We offer an open-source library for \MethodName, facilitating its easy integration with PyTorch frameworks.\footnote{https://github.com/sohail2810/LayerCollapse}
\end{itemize}

\section{Background}

Multi-Layer Perceptrons (MLPs) are foundational elements in modern deep learning architectures, integral to tasks such as classification, regression, and sequence modeling. Structurally, an MLP comprises fully connected layers where each neuron in a layer connects to every neuron in the subsequent layer. Originally developed as one of the first neural network models \cite{haykin1994neural}, MLPs remain essential in contemporary architectures like transformers and vision-based networks \cite{vaswani2017attention, dosovitskiy2020image} due to their capacity to model complex relationships and approximate intricate functions.

\subsection{Role and Structure of MLPs}

MLPs consist of sequences of linear transformations interspersed with non-linear activation functions, where the linear transformation can be represented as $\mathbf{y} = \mathbf{W} \mathbf{x} + \mathbf{b}$. Without the non-linear functions, consecutive linear transformations would collapse into a single transformation, limiting the model's ability to learn non-linear relationships \cite{goodfellow2016deep}. Popular activation functions in MLPs include ReLU, parametric ReLU (PReLU), GeLU, ELU, Sigmoid, and Tanh, with ReLU widely used for its computational efficiency and mitigation of the vanishing gradient problem \cite{nair2010rectified, glorot2011deep}.

\subsection{MLPs in Modern Architectures}

MLPs are pivotal in advanced architectures like transformers, which utilize MLP layers following self-attention mechanisms to enhance representational capacity \cite{vaswani2017attention}. In Vision Transformers (ViTs), MLPs process image patch embeddings, enabling spatial and feature learning from image data \cite{dosovitskiy2020image}. Architectures like MLP-Mixers further showcase the versatility of MLPs by leveraging them for token and channel mixing, demonstrating that MLPs alone can achieve competitive performance in vision tasks \cite{tolstikhin2021mlp}.

\subsection{Importance of Non-Linear Activations}

The inclusion of non-linear activation functions between linear layers is essential for MLPs to capture complex, non-linear mappings. Activation functions like ReLU introduce discontinuities that prevent the model from being purely linear, while bounded functions like Sigmoid and Tanh enable stability in certain applications despite their tendency to saturate \cite{glorot2011deep}. By breaking up linear layers, non-linearities enable MLPs to approximate complex functions effectively, a quality exploited in diverse applications.

\subsection{Applications and Challenges}

MLPs are critical in classification heads for converting high-dimensional representations to class scores in tasks such as image classification and sentiment analysis \cite{ILSVRC15}. In regression tasks, MLPs predict continuous values and are widely used in financial forecasting and scientific applications. Despite these strengths, the dense nature of MLPs contributes significantly to the computational overhead in models, spurring the need for optimization techniques like pruning, quantization, and activation modulation \cite{fan2019reducing, hubara2017quantized}. Activation modulation, in particular, can simplify models by modulating non-linear activations toward linearity, thus reducing computational demands without heavily impacting model expressiveness.




\section{Methodology}
\label{sec:methodology}

This section presents \MethodName, a novel regularization technique designed to adaptively compress the model by reducing its depth, possibly improving generalization in the process. In \MethodName, we employ compression-aware training. This approach enables the extraction of a shallower model directly, without the need for finetuning.

As illustrated in Figure \ref{fig:main_idea}, LayerCollapse involves regularizing the complexity of a layer by linearizing its activation function when training or fine-tuning. This linearization enables the merging of adjacent linear layers into a single layer, thus effectively reducing the model's depth. The resulting model has lower storage, and computation demands without the need for software/hardware support.

To gain intuition on the effect that the non-linearities have on the model's representation power and complexity, we present a toy scenario. Consider a 2-layer MLP with a hidden dimension of 500 and input and output dimensions of 1, employing a ReLU non-linearity. This network can be used to fit a polynomial function $f(x)=(x-3)(5x+2)(x+1)$ with samples having Gaussian noise. We apply gradient descent to fit the MLP to this function, aiming to uncover the underlying function $f(x)$. The results of this experiment are shown in Figure \ref{fig:1d_demo}. In this demo, the regular model is described above. The $n\%$ linear model is a leaky ReLU with a negative slope of $n\%$. We observe that at $n=0.8$, the complexity of the 2-layer network is reduced such that it does not overfit to noise, closely approximating the underlying function. Conversely, at $n=0.98$, the reduced complexity of the model leads to underfitting. Setting $n=100\%$ results in a linear function, $Activation(x) = x$, effectively simplifying the 2-layer MLP to just a linear transformation since the product of the matrices $L_1^{1\times m}$ and $L_2^{m\times1}$ results in a scalar. Here, $L_1 \in \mathbb{R}^{1 \times m}$ and $L_2 \in \mathbb{R}^{m \times 1}$ are the weight matrices of the first and second layers.

Following the intuition from this experiment, we can see the effect of the negative slope on the model's expressiveness. We design a regularizer that adaptively controls the negative slope of the MLP, preventing the model from overfitting to the noise. If for any layer we find that the negative slope is one, we can remove that layer from the network following the procedure in Algorithm \ref{alg:collapse} without losing any performance. The resulting network will have a shallower depth, fewer parameters, and lower resource requirements for execution.

\subsection{Formal Definition}
\label{subsec:formal_def}
We define a Collapsible MLP as: Two consecutive fully connected layers utilizing a Parametric Rectified Linear Unit (PReLU) \cite{he2015delving} as the activation. The Parametric Rectified Linear Unit (PReLU) can be expressed as:
\begin{equation}
\label{eq:relu}
PReLU_\alpha(x) = \max(0, x) + \alpha\min(0, x)
\end{equation}
where $\alpha$ represents the slope of the PReLU function when x is negative.
Output of a basic two-layer collapsible MLP is described as:
\begin{equation}
Y_\alpha = W_2(PReLU_\alpha(W_1X + b_1)) + b_2
\end{equation}
where, $X$ is the input random variable and $Y$ is the output random variable. $W_1$ and $W_2$ are the weight matrices and $b_1$ and $b_2$ are the bias vectors for the first and second fully connected layers, respectively.
$\alpha$ can be used to make the activation linear.
We define $Y_{linear}$ as:
\begin{equation}
    Y_{linear}=W_2W_1X+W_2b_1+b_2
\end{equation}
We define the non-linearity error as the squared difference of $Y_\alpha$ and $Y_{linear}$.
In Appendix \ref{sup:mse}, we show that $\forall \delta \in [0,1] $
\begin{equation}
\label{eq:p_error}
    P\left\{ |Y_{\text{linear}} - Y_\alpha|^2 \leq C\times(1-\alpha)^2 \right\} > 1 - \delta
\end{equation}
where, 
\begin{equation}
\begin{aligned}
     &x^\delta\equiv \inf\left\{x\in\mathcal{R}^n\mid P\left\{ |X|>|x| \right\}<\delta\right\}\\
     &C = \sigma_{max}(W_2W_1)^2|x^\delta|^2+|W_2b_1|^2 \\
\end{aligned}
\end{equation}
where, $\sigma_{max}(W_2W_1)$ is the largest singular value of $W_2W_1$
By choosing $\delta$ arbitrarily small, probability of error is bounded proportional to $(1-\alpha)^2$ with probability one.
We incorporating $(1-\alpha)^2$ into our loss function to optimize for the non-linearity loss as well as the label cross-entropy loss.
This loss term serves to align $Y_\alpha$ with $Y_{linear}$, compressing the model into a shallower model as demonstrated in Figure \ref{fig:main_idea}. In simpler terms, as $\alpha$ approaches one, substituting $Y_{linear}$ with $Y_\alpha$ incurs a bounded error probability, the error probability is proportional to $(1-\alpha)^2$ times a constant.

In practical applications, \textbf{Dropout} \cite{srivastava2014dropout} and \textbf{BatchNormalization} \cite{ioffe2015batch} are frequently used between the layers of a network. Dropout works by temporarily deactivating random neurons during training, which helps spread the learning across different neurons to avoid over-reliance on any particular subset of them. It is important to note that Dropout is not active during model deployment, meaning it behaves as if it is not there. This characteristic means that Dropout doesn't affect the layer collapsing process and can be left out of consideration.\\

BatchNormalization is a process described by the formula:
\begin{equation}
Y = \frac{X - E[X]}{\sqrt{Var[X]}} \times \gamma + \beta
\end{equation}
where, $X$ and $Y$ are the input and output random variables. $\gamma$ and $\beta$ are learnable parameters, and $E[X]$ and $Var[X]$ are the mean and variance of the current batch. 

Algorithm \ref{alg:collapse} outlines the specific steps to be taken for adjusting weights and biases when BatchNormalization is present.

\subsection{Compression Ratio}
Compression ratio for the collapsed layer can be calculated as:
\begin{equation}
\label{eq:gain}
CR = 1-\frac{n_{in}\times n_{out}}{n_{hidden}\times(n_{in}+n_{out})}
\end{equation}
where $CR$ is the compression ratio, $n_{in}$, $n_{hidden}$, and $n_{out}$ denote the input size, hidden size, and output size of the 2-layer MLP, respectively.

Noteworthy is that the compression ratio can be negative. Positive compression ratio only happens when $n_{hidden}$ is larger than the product of $n_{in}$ and $n_{out}$ divided by their sum. This means that MLPs designed with a ``bottleneck" structure (where $n_{hidden}$ is relatively small) may not see a benefit from \MethodName. However, most modern MLP designs tend to expand the layer size (``widening" structure), and are likely to have significant compression ratios.

Convolutional layers play a pivotal role in numerous neural network architectures. Essentially,  convolution can be formulated as sparse fully connected layers with some redundancies. Their nature as linear transforms permits representation via circulant matrices. A circulant matrix is a matrix where the row vectors have the same elements and each row vector is rotated one element to the right relative to the preceding row vector. The product of two circulant matrices is also a circulant matrix. So collapsing two convolution layers will result in another convolution layer.
The kernel size of the resulting convolution layer, denoted as
$k$, is calculated as $k=k_1+k_2-1$ where, $k_1$ and $k_2$ are the kernel sizes of the original convolution layers. 
Compression gain for convolution layers can be calculated as:
\begin{equation}
CR_{conv} = 1 - \frac{k_1^2 \times c_{in} \times c_{hidden} + k_2^2 \times c_{hidden} \times c_{out}}{(k_1 + k_2 - 1)^2 \times c_{in} \times c_{out}}
\end{equation}
where, $c_{in}$, $c_{hidden}$, and $k_1$ correspond to the input channels, output channels, and the kernel size of the initial convolution layer, respectively. $c_{hidden}$, $c_{out}$ and $k_2$ are the parameters of the subsequent convolution layer.
However, a crucial point to consider when applying \MethodName to common CNN architectures, such as ResNet, is that the compression ratio is negative. This is due to the fact that a positive compression ratio requires for $c_{hidden}$ to be significantly larger than both $c_{in}$ and $c_{out}$. This observation highlights the need for careful consideration when attempting to collapse layers in such architectures. 

\subsection{Prior Distribution Analysis of Regularization}
\label{subsec:prior_dist_reg}

We extend the use of the parametric ReLU (PReLU) function to develop a regularization method based on the maximum a posteriori (MAP) framework \cite{degroot2005optimal}. 
Aligning with Occam's razor \cite{schaffer2015not}, which favors simpler explanations, we propose to adopt a normal distribution with a mean of 1 as the prior distribution for $\alpha$.
This setting effectively limits the two layers to a linear space, reducing their functional complexity.
This choice is backed by empirical results given in section \ref{sec:experiment}. An intuition for this choice is presented in figure \ref{fig:1d_demo}
MAP framework suggests adding negative log likelihood of the prior distribution of parameters to the loss function. This is referred to as the prior loss or regularization loss. Following this principal the added regularization term is:

\begin{equation}
\label{eq:regularizer}
    \mathcal{L}_{\text{reg}} = lc \times (1 - \alpha)^2,\text{}
\end{equation}
where $lc$ is the regularization constant. Notice this formulation aligns perfectly to our intuition from equation \eqref{eq:p_error}.


\begin{algorithm}[H]
\caption{Layer Collapse}
\label{alg:collapse}
\begin{algorithmic}
    \FUNCTION{Collapse($W_1, b_1, W_2, b_2, \gamma, \beta, \bar{\mu}, \bar{\sigma}, \alpha, \tau$)}
        \IF{$\alpha \leq \tau$}
            \STATE $s_{BN} \gets \frac{\gamma}{\sqrt{\bar{\sigma}}}$
            \STATE $W_{\text{new}} \gets s_{BN} \times W_2 \cdot W_1$
            \STATE $b_{\text{new}} \gets W_2 \cdot (s_{BN} \times (b_1 - \bar{\mu}) + \beta) + b_2$
            \STATE \textbf{return} $W_{\text{new}}, b_{\text{new}}$
        \ELSE
            \STATE \textbf{return} un-collapsible
        \ENDIF
    \ENDFUNCTION
\end{algorithmic}
\end{algorithm}


\begin{table*}[t]
    \centering
        \caption{Training performance on Wikitext-103 dataset. For base variants the modified layer is the last layer i.e. layer 12. In the large variant layers 11, 16, and 32 are selected for compression.}
        \label{tab:ppl}
        \begin{tabular}{lclc}
            \toprule
            Model & Layers & Params (M) & PPL \\
            \midrule
            GPT-2 \cite{radford2019language} & 0 & 124.4 & \textbf{18.87} \\
            GPT-2 + LayerDrop \cite{fan2019reducing} & 1 & 119.9(-4\%) & 19.29 \\
            GPT-2 + \MethodName (ours) & 1 & 120.3(-3\%) & 19.09 \\
            \midrule
            GPT-2 Large \cite{radford2019language} & 0 & 774.0 & 17.65 \\
            GPT-2 Large + LayerDrop \cite{fan2019reducing} & 3 & 734.7(-5\%) & 15.61 \\
            GPT-2 Large + \MethodName (ours) & 3 & 739.6(-4\%) & \textbf{15.17} \\
            \bottomrule
        \end{tabular}
\end{table*}

\subsection{Alternative Activation Functions}
Thus far, we have focused on the ReLU function as a primary non-linear activation function. ReLU’s simplicity and efficiency have had a significant impact on deep learning since its introduction. Recently, several ReLU variants, including ELU \cite{clevert2015fast}, GeLU \cite{hendrycks2016gaussian}, and SiLU \cite{elfwing2018sigmoid}, have gained popularity due to their enhanced flexibility.

We propose a general approach to smoothly linearize ReLU-family activations. For functions like ELU, which are linear in the positive region and non-linear in the negative, we incorporate a linear term, $y = x$, into the negative region. This term is blended with the original function using a weighted sum, where the weight parameter is denoted by $\alpha$ as defined in section \ref{subsec:formal_def}. For example, the modified ELU function can be written as:
\[
\text{ELU}^\beta(x) = 
\begin{cases} 
      x & \text{if } x > 0 \\
      \alpha x + (1 - \alpha) \beta (\exp(x) - 1) & \text{if } x \leq 0 
\end{cases}
\]
where $\beta$ is the ELU parameter. When $\alpha = 1$, ELU transitions to a linear function. To ensure $\alpha$ remains within $[0, 1]$ during training, we apply a sigmoid to it.

For one-region activations like GeLU and SiLU, which follow the form $f(x) = x \times h(x)$ with $h(x) \rightarrow 1$ as $x \rightarrow \infty$ and $h(x) \rightarrow 0$ as $x \rightarrow -\infty$, we adjust using $\alpha$ in the form $f(x) = x \times (h(x) + \alpha (1 - h(x)))$, where $\alpha \in [0,1]$. This leads to parametric versions:
\[
\text{GeLU}^\alpha(x) = x(\Phi(x) + \alpha (1 - \Phi(x)))
\]
where $\Phi(x)$ is the Gaussian cumulative distribution function, and for SiLU:
\[
\text{SiLU}^\alpha(x) = x(\sigma(x) + \alpha (1 - \sigma(x)))
\]
where $\sigma(x) = \frac{e^x}{1 + e^x}$ is the sigmoid function.

\section{Experiments}
\label{sec:experiment}

We conduct a comprehensive evaluation of \MethodName's accuracy and efficiency. For accuracy, we demonstrate \MethodName's capability on NLP and vision tasks, showcasing its effectiveness as both a regularizer and a compression method. 

Experiments presented in Section \ref{sec:experiment_reg} illustrate \MethodName's dual role not only as a compression tool for pre-trained networks but also in enhancing model generalization. We compare \MethodName against a modified version of \textit{LayerDrop}, a structured pruning technique, highlighting the distinct advantages of our approach.

Detailed in Section \ref{subsec:compression}, we finetune models with \MethodName for regularization followed by compression. Additionally, we compare \MethodName against Knowledge Distillation, as discussed in \cite{hinton2015distilling}, focusing on the computational resources required for each method, despite targeting the same network architectures.

Finally, we test \MethodName as a regularizer by training a VGG network from scratch, presenting two scenarios: using \MethodName and not using it, to evaluate its impact on model performance and efficiency.

\begin{table*}[t]
    \centering
        \caption{ 
        Post-training compression GLUE test results illustrating accuracy and achieved compression. For QQP, MRPC average of F1 and accuracy are reported, for STS-B average of Spearman and Pearson is reported.}
        \label{tab:compression_glue}
        \begin{tabular}{p{0.7cm}clcccccccc|c}
            \toprule
            Model & Layers & Params (M) & MNLI(m/mm) & SST-2 & STS-B & RTE & QNLI & QQP & MRPC & CoLA & Average \\
            \midrule
            \multirow{2}{*}{Bert Base} 
                 & 0 & 109.1  & 83.5/83.7 & 92.7 & 86.4 & 65.7 & 91.0 & 89.4 & 85.8 & 61.0 & 81.9\\
            & 4 & 92.5(-15\%) & 82.6/82.8 & 91.2 & 87.3 & 61.4 & 89.8 & 86.3 & 88.4 & 59.6 & 80.8\\
            \cmidrule{2-12}

            \multirow{2}{*}{Bert Large} 
                 & 0 & 334.9  & 86.3/\textbf{86.1} & \textbf{93.5} & 87.7 & 66.4 & 89.3 & \textbf{89.9} & 85.8 & 62.6 & 82.7\\
            & 4 & 309.7(-8\%) & \textbf{86.6}/85.5 & 93.0 & \textbf{89.5} & \textbf{71.5} & \textbf{91.4} & 89.4 & \textbf{87.5} & \textbf{62.6} & \textbf{83.8}\\
            \bottomrule
        \end{tabular}
\end{table*}

\subsection{Regularization Performance of \MethodName}
\label{sec:experiment_reg}

\subsubsection{Pre-Trained Models}
\textbf{Setup.}
In our experiments, we utilize the base and large versions of the GPT-2 causal language model \cite{radford2019language}, with a fixed context size of 128 tokens. Our layer selection methodology, inspired by \cite{fan2019reducing}, involves initially collapsing each layer, using the algorithm presented in \ref{alg:collapse}, in a one-shot manner, followed by an evaluation to identify layers most tolerant to compression. We evaluate \MethodName on the widely used perplexity metric on the test set of the Wikitext-103 dataset \cite{merity2016pointer}.

We conduct training for $3k$ steps, maintaining a batch size of 128 across all experiments. The learning rate scheduling is managed by LiveTune \cite{shabgahi2023livetune}, starting from an initial rate of $0.001$. Optimization is performed using a modified version of the Adam optimizer \cite{kingma2014adam}, with comprehensive details about this optimizer available in Appendix \ref{sup:adam}. An exhaustive list of experiment details can be found in Appendix \ref{sup:experiments}. The comparison was done with a modified version of \textit{LayerDrop} \cite{fan2019reducing}. This modification, drawing inspiration from \cite{savarese2017residual}, employs gates for selected layers. These gates use a sigmoid function to dynamically alternate between activation and deactivation, enhancing training stability.

\textbf{Results.} As shown in Table \ref{tab:ppl}, \MethodName surpasses \textit{LayerDrop} in terms of final perplexity score. Notably, the large variant of GPT-2 tends to overfit, resulting in higher Perplexity compared to both regularized approaches. However, \textit{LayerDrop} achieves greater compression gain.

\subsubsection{Trained from Scratch.}
\label{subsec:ablation}
We provide an in-depth analysis of how our regularization approach impacts both the generalization and compression abilities in training image classifiers from scratch.

\noindent\textbf{Dataset. }Our study utilizes the CIFAR100 \cite{han2015learning} and TinyImageNet \cite{le2015tiny} datasets, which are mid-sized and rich in diversity. The CIFAR100 dataset includes 60,000 images across 100 classes, and TinyImageNet, a subset of the larger ImageNet, comprises 100,000 images across 200 classes.\\
\noindent\textbf{Implementation. } The experiments are conducted using the VGG architecture because of its effective learning capabilities without pre-training and competitive performance in benchmarks. For consistency, all experiments utilize the same setup and hyper parameters. Note that the goal of this study is not to show state of the art performance using our proposed regularization, but to show its effectiveness in reducing over-fitting and final model size. In all experiments, traditional regularizers, such as data augmentation (rotation, random crop, and flipping), dropout\cite{do2021maxdropout}, and batch normalization\cite{ioffe2015batch}, have been used. This is to show the added contribution of \ref{eq:regularizer}. Training is done using Stochastic Gradient Descent (SGD) with momentum a learning rate scheduler, trained for a total of 400 epochs per experiment. The regularization is applied to 40\% of the layers, starting with the latest layers, with the strength of 0.05 across all experiments. extensive experiment details can be found in Appendix \ref{sup:sub:ablation}.\\

\begin{figure}[]
  \centering
   \includegraphics[width=1\linewidth]{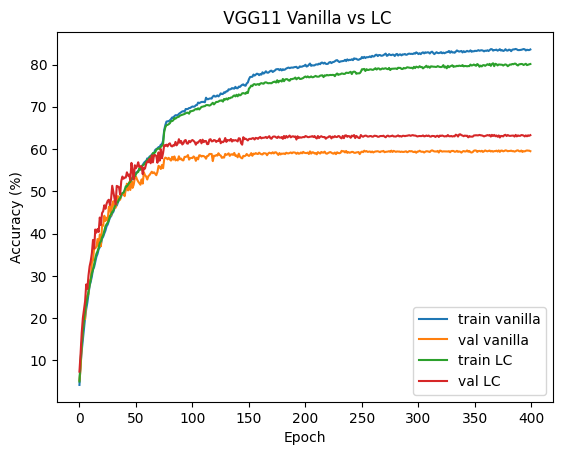}

   \caption{Evaluating the impact of LayerCollapse regularization on CIFAR100, we compare the top-1 accuracy of VGG11 both with and without this regularization technique.}
   \label{fig:reg_vanilla_vs_lc}
\end{figure}

\noindent\textbf{Discussion. }
The results, documented in Table \ref{tab:accuracy_comparison}, demonstrate the benefits of incorporating compression aware regularization. We observe an enhancement in top-1 accuracy by as much as 2\% with 50\% reduction in model size. The learning curve presented in Figure \ref{fig:reg_vanilla_vs_lc} suggests a narrowing gap between training and validation accuracy, indicating a reduction in overfitting and highlighting the effectiveness of our regularization strategy in improving generalization.

\begin{table}[t]
\centering
\caption{Number of parameters after Collapse and Top-1 Accuracy of various models with and without LC on CIFAR-100 and Tiny ImageNet.}
\label{tab:accuracy_comparison}
\begin{tabular}{
  @{}l 
  S[table-format=2.1] 
  S[table-format=2.1] 
  S[table-format=2.1] 
  S[table-format=2.1] 
  @{}
}
\toprule
\multirow{2}{*}{Model} & \multicolumn{2}{c}{CIFAR-100} & \multicolumn{2}{c}{Tiny ImageNet} \\
\cmidrule(lr){2-3} \cmidrule(lr){4-5}
& {\#Param} & {Top-1(\%)} & {\#Param} & {Top-1(\%)} \\
\midrule
VGG11      & 11.7M & 59.7 & 18.4M & 60.2 \\
VGG11(LC) & 9.3M & 63.5 & 9.6M & 61.7 \\
\addlinespace
VGG13      & 11.9M & 66.1 & 18.6M & 50.3 \\
VGG13(LC) & 9.5M & 67.8 & 9.8M & 52.2 \\
\addlinespace
VGG16      & 17.2M & 63.3 & 23.9M & 48.9 \\
VGG16(LC) & 14.8M & 64.1 & 15.1M & 49.7 \\
\addlinespace
VGG19      & 22.5M & 62.5 & 29.6M & 46.8 \\
VGG19(LC) & 20.1M & 63.8 & 20.4M & 48.2 \\
\bottomrule
\end{tabular}
\end{table}


\subsection{\MethodName for Compression}
\label{subsec:compression}
We apply \MethodName to compress targeted layers in pretrained models in both computer vision and NLP tasks. To utilize \MethodName as compression for pre-trained models, a few rounds of fine-tuning with the \MethodName regularizer are required to ensure the linearization does not significantly impact accuracy. The process involves a few training steps with the regularization loss defined in Equation \ref{eq:regularizer}. In all cases, we set the slope threshold for layer collapse at $0.0001$. It's important to note that we report results directly after layer collapse, without the need for additional fine-tuning.

For architectures such as Bert, ViT, and MLP-Mixer, we substitute GeLU activations in the compressed layers with ReLU, as GeLU is a smoothed ReLU and this change has been found to be insignificant in our tests. Our compression approach focuses on the model's final layers, driven by the insight that these layers handle more abstract data representations, rendering them more suitable for compression. The compression is carried out  one layer at a time. For detailed implementations notes, please refer to Appendix \ref{sup:experiments}.

\subsubsection{Performance on GLUE}
\label{subsubsec:glue}
\textbf{Setup.}
We assess \MethodName's compression effectiveness on pre-trained Bert-Large and Bert-Base \cite{devlin2018bert} using the GLUE benchmark \cite{wang2018glue}.The \textit{Average} column is slightly different than GLUE score since we exclude WNLI as discussed in \cite{devlin2018bert}.

\textbf{Results.} 
As illustrated in table \ref{tab:compression_glue} \MethodName yields $15\%$ compression gain for Bert base with about $1\%$ score reduction on the GLUE benchmark. As evident the larger variant of Bert is prone to overfit thus \MethodName regularization improves the score in many tasks. As illustrated \MethodName provided better scores than the baseline in \cite{devlin2018bert}. This is while the resulting model is $8\%$ smaller in terms of parameter count.

\begin{table*}[t]
    \centering
        \caption{ 
        Post-training compression results, showcasing accuracy and compression attained 
        via \MethodName across different layers in relevant architectures.}
        \label{tab:compression_imagenet}
        \begin{tabular}{lcllc}
            \toprule
            Model & Layers & Params (M)  & MACs (G) & Top1 Acc. (\%) \\
            \midrule
            \multirow{2}{*}{ViT T/16} & 0 & 5.72 & 1.08 & 75.35 \\
            & 3 & 4.94(-14\%) & 0.91(-16\%) & 74.42 \\
            \cmidrule{2-5}
            \multirow{2}{*}{ViT S/16} & 0 & 22.05 & 4.24 & 81.38 \\
            & 2 & 19.98(-9\%) & 3.84(-9\%) & 79.74 \\
            \cmidrule{2-5}
            \multirow{2}{*}{ViT B/16} & 0 & 86.57  & 16.85 & 81.43 \\
            & 2 & 78.30(-10\%) & 15.23(-10\%) & 80.31 \\
            \cmidrule{2-5}
            \multirow{2}{*}{ViT L/16} & 0 & 304.33  & 59.66 & 84.80 \\
            & 2 & 289.64(-5\%) & 56.77(-5\%) & 84.22 \\
            \midrule
            \multirow{2}{*}{Mixer B/16} & 0 & 59.88 & 12.61 & 76.47 \\
            & 4 & 51.39(-14\%) & 10.81(-14\%) & 75.67  \\
            \midrule
            \multirow{2}{*}{VGG19} & 0 & 143.67 & 19.65 & 74.21 \\
            & 2 & 45.11(-67\%) & 19.55(-1\%) & 71.22 \\
            \cmidrule{2-5}
            \multirow{2}{*}{VGG11} & 0 & 132.86 & 7.62 & 70.37 \\
            & 2 & 34.31(-74\%)& 7.52(-1\%) & 66.36  \\
            \bottomrule
        \end{tabular}
\end{table*}

\subsubsection{Performance on ImageNet-1K}
\label{subsubsec:imagenet}
\textbf{Setup.}
We apply \MethodName across a diverse set of vision models, including CNNs, MLP-Mixers \cite{tolstikhin2021mlp}, and Vision Transformers \cite{dosovitskiy2020image}, utilizing the ImageNet-1K benchmark \cite{ILSVRC15} for evaluation. Pretrained models are sourced from the timm library \cite{rw2019timm}. Optimization is performed using SGD with momentum, with a learning rate set to $5e^{-4}$. 
We employ the \textit{Self Distillation} method \cite{fan2019reducing}. The loss function is comprised of three parts, cross entropy between labels and result, KL divergence between the output and original model output as the teacher and the LayerCollapse regularization term given in equation \eqref{eq:regularizer} with $lc=0.2$. This process is capped at 20 epochs. Details are enumerated in Appendix \ref{sup:experiments}.

\begin{figure}[]
  \centering
   \includegraphics[width=1.0\linewidth]{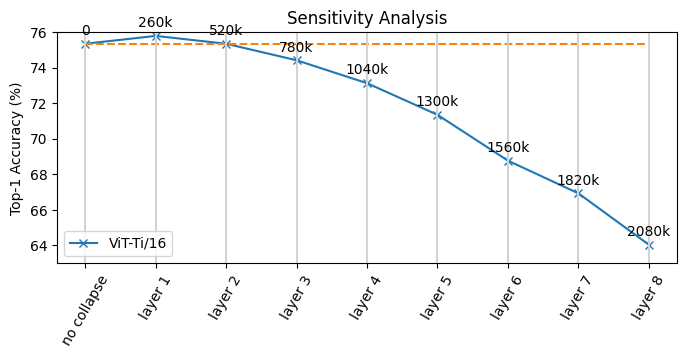}

   \caption{Layer-wise collapse accuracy and parameter reduction analysis for ViT-T/16.}
   \label{fig:sensitivity}
\end{figure}

\textbf{Sensitivity Analysis. } To quantify the robustness of LayerCollapse, we conduct a layer-by-layer sensitivity analysis on the Vision Transformer T/16 model, as shown in Figure \ref{fig:sensitivity}. This removal of layers provides insight into the impact of incremental compression on the model's accuracy. We observe an initial increase in accuracy upon the collapse of the first layer, attributed to an improvement in generalization compared to the original model.

\textbf{Results.}
Table \ref{tab:compression_imagenet} illustrates that LayerCollapse achieves a substantial reduction in model size-up to $74\%$ in VGG11—with minimal accuracy degradation. Notably, models with heavier classification heads, such as VGG, can benefit significantly from our method. In all ViT variants we get a consistent reduction of over $10\%$ in size and complexity with less than a 1\% accuracy loss. In the Mixer architecture, 2 types of MLP layers known as MLP-token and MLP Channels exist. MLP-Channel has a significantly bigger size. We found that MLP-tokens can be collapsed in about an epoch of computation whilst MLP-channels require more effort. In the Mixer experiment, we collapsed the layers in pairs to ensure MLP-Channels are affected. In the Mixer architecture we get $14\%$ reduction in size with less than a $1\%$ accuracy degradation.

\subsection{LayerCollapse vs Knowledge Distillation}
\label{subsec:lc_vs_kd}

\textbf{Setup.} 
Our evaluation framework utilizes batch-normalized VGG variants from the torchvision repository \cite{torchvision2016}. We use the ImageNet-1K \cite{ILSVRC15}  dataset, providing a substantial dataset for training and validation. These variants are chosen for their prevalent use in compression studies. To facilitate a fair comparison between LayerCollapse and Knowledge Distillation, we adopt the same target architecture for both methods. The rest of the training is set up identical to section \ref{subsec:compression} for both LayerCollapse and Knowledge distillation.

\textbf{Comparison. } The models are rigorously compared across several dimensions: training cost, the number of training epochs required, and the accuracy metrics—top-1 and top-5—on the validation set.
The computational costs are quantified on one Nvidia RTX A6000 GPU.


\vspace{50px}
\textbf{Results.} 
Referencing table \ref{tab:distillation_comparison}, LayerCollapse consistently achieves over 10\% higher accuracy while incurring only 30\% of the computational cost compared to Knowledge Distillation. This trend is more pronounced in larger variants, which exhibit greater compression potential. Showing up to 15\% accuracy gain with 50\% of tuning cost. It is noteworthy that for models necessitating pre-training, such as MLP-Mixer and ViT, the absence of pre-training data and the high computational demands often make Knowledge Distillation impractical. However, we recognize the inherent limitations of \MethodName in terms of the diversity of achievable target model architectures. \MethodName can only remove fully connected layers while model distillation can change the architecture completely.




\section{Discussion}

In this study, we presented \MethodName, a novel approach for compressing neural networks through regularization techniques that strategically merge linear layer pairs, effectively reducing both network depth and parameter count. This approach specifically encourages the formation of shallower network structures, contributing to computational efficiency. Evaluations show that \MethodName achieves better performance in regularization and compression than other structural modifications. Compared to Knowledge Distillation, \MethodName delivers over 10\% higher performance while using 80\% less computational resources. However, the method’s strengths primarily benefit architectures featuring MLPs with widening designs, and it does not substantially improve compression for current CNN models.

\begin{table*}[t]
    \centering
    \caption{An analysis of \MethodName (LC) and Knowledge Distillation (KD) methods applied to VGG variants.}
    \label{tab:distillation_comparison}
    \begin{tabular}{lclccc}
        \toprule
        Model & Method & Top1 Acc(\%) & No. of Parameters (M) & Cost & Epochs \\
        \midrule
        \multirow{2}{*}{vgg19} & LC & 71.22 & 45.11 & 11h & 5 \\
                               & KD & 63.42 & 45.11 & 51h & 23 \\
        \cmidrule{2-6}
        \multirow{2}{*}{vgg16} & LC & 70.01 & 39.80 & 14h & 7 \\
                               & KD & 59.63 & 39.80 & 31h & 16 \\
        \cmidrule{2-6}
        \multirow{2}{*}{vgg13} & LC & 67.21 & 34.49 & 16h & 10 \\
                               & KD & 58.26 & 34.49 & 45h & 28 \\
        \cmidrule{2-6}
        \multirow{2}{*}{vgg11} & LC & 66.36 & 34.31 & 20h & 20 \\
                               & KD & 58.64 & 34.31 & 42h & 42 \\
        \bottomrule
    \end{tabular}
\end{table*}

\section{Related Work}
\label{sec:Relatedwork}

In overparameterized neural networks, two major concerns are generalizability and computational requirements. Compression techniques aim to address the second issue by reducing computational demands while maintaining model precision. The predominant methods for model compression include pruning \cite{han2015learning}, quantization \cite{hubara2017quantized}, and knowledge distillation \cite{hinton2015distilling}, each contributing uniquely to model optimization \cite{gupta2022compression}. 
Quantization reduces the number of bits needed to store weights, though it requires hardware support to be effective. Pruning strategies fall into two categories, structured and unstructured. Unstructured pruning reduces storage needs by utilizing sparse representations, but high prune rates and hardware support are necessary. Structured pruning \cite{fan2019reducing} \cite{lemaire2019structured} \cite{fu2022depthshrinker} changes the model architecture, obviating the need for additional hardware or software support. Knowledge distillation changes the model's architecture by transferring knowledge from a large model to a smaller one. The problem with knowledge distillation is the immense computational demands of training the target architecture.
\MethodName, similar to structured pruning and knowledge distillation, modifies the neural network architecture to achieve compression without requiring further software or hardware optimization. Our method compares favorably to LayerDrop \cite{fan2019reducing} and knowledge distillation \cite{hinton2015distilling}, as shown in Section \ref{sec:experiment}.
Pruning and quantization can be iterative or one-shot. Iterative methods compress the model in steps with finetuning, while one-shot methods incorporate regularization during training to enable compression without further finetuning. Quantization-aware training (QAT) \cite{nguyen2020quantization} and pruning-aware training (PAT) \cite{lemaire2019structured}\cite{jiang2023pruning}\cite{wen2016learning} \cite{alvarez2016learning} are common practices to mitigate accuracy loss. LayerDrop is a notable PAT that improves training results as a regularizer \cite{fan2019reducing}. In Section \ref{sec:experiment_reg}, we evaluate the regularization effect of \MethodName against LayerDrop.

Explicit regularization methods, such as Dropout \cite{do2021maxdropout}, \cite{ghiasi2018dropblock}, \cite{srivastava2014dropout}, and weight decay, follow the principle that simpler models are more robust \cite{schaffer2015not} and less prone to overfitting. Following the same principle, we adjust the model's structure with \MethodName to prevent overfitting. However, this principle has been shown to be inaccurate in recent over-parameterized models when employed with immense amounts of data.

Previous work on private inference has explored eliminating non-linearities in neural networks for computational efficiency of private inference. DeepReDuce \cite{jha2021deepreduce} and Selective Network Linearization (SNL) \cite{cho2022selective} adopt regularization strategies for linearization. Unlike SNL, \MethodName introduces a theoretical regularizer and demonstrates significant efficiency gains across computer vision and natural language processing domains.

\section{Conclusion and Limitations}

\MethodName advances the development of efficient neural networks by optimizing performance alongside a reduction in model size and computational needs. However, the utility of \MethodName is somewhat restricted as it primarily compresses fully connected layers, which limits its effectiveness in CNN architectures. Moreover, this method necessitates either finetuning or pretraining to prepare the model for compression, and it also demands precise adjustment of the hyperparameter $\alpha$ to prevent a decline in performance and accuracy. These limitations highlight areas for further research to broaden the application of \MethodName to a more diverse array of network structures and model types.


\bibliography{example_paper}
\bibliographystyle{mlsys2025}

\medskip

\newpage

\appendix

\section{Proof of Equation \eqref{eq:p_error}}
\label{sup:mse}
Given $Y_{linear}=W_2W_1X+W_2b_1+b_2$ and $Y_\alpha=W_2PReLU_\alpha(W_1X+b_1)+b_2$ we show that $\forall \delta > 0$
\begin{equation}
\label{sup:eq:p_error}
    P\left\{ |Y_{\text{linear}} - Y_\alpha|^2 \leq C\times(1-\alpha)^2 \right\} > 1 - \delta
\end{equation}
where, 
\begin{equation}
\label{sup:eq:xdelta}
\begin{aligned}
     &x^\delta\equiv \inf\left\{x\in\mathcal{R}^n\mid P\left\{ |X|>|x| \right\}<\delta\right\}\\
     &C = \sigma_{max}(W_2W_1)^2|x^\delta|^2+|W_2b_1|^2 \\
\end{aligned}
\end{equation}
where, $\sigma_{max}(W_2W_1)$ is the largest singular value of $W_2W_1$.

We define indicator random variable $I$ as:
\begin{equation}
\label{sup:eq:I}
I \equiv
    \begin{cases}
        1 & W_1X+b_1\geq 0\\
        0 & otherwise
    \end{cases}
\end{equation}

Writing $Y_\alpha$ using \eqref{sup:eq:I} we have:
\begin{equation}
    Y_\alpha = 
        IY_{linear} + (1-I)(\alpha W_2W_1X+\alpha W_2b_1 + b_2)
\end{equation}
Using the definition of $Y_{linear}$, the squared distance of $Y_{linear}$ and $Y_\alpha$ can be written as:
\begin{equation}
    |Y_{\text{linear}} - Y_\alpha|^2 = (1-I)^2(1-\alpha)^2(W_2W_1X+W_2b_1)^2
\end{equation}
Since $(1-I)^2$ is either $0$ or $1$ we have:
\begin{equation}
    |Y_{\text{linear}} - Y_\alpha|^2 \leq (1-\alpha)^2(W_2W_1X+W_2b_1)^2
\end{equation}
Using the triangle inequality we can write:
\begin{equation}
    |Y_{\text{linear}} - Y_\alpha|^2 \leq (1-\alpha)^2(|W_2W_1X|^2+|W_2b_1|^2)
\end{equation}
It is known that:
\begin{equation}
\label{sup:eq:singular}
    |W_2W_1X|^2 \leq \sigma(W_2W_1)^2|X|^2,
\end{equation}
where, $\sigma(W_2W_1)$ is the largest singular value of the $W_2W_1$ matrix.
Using \eqref{sup:eq:singular} we have:
\begin{equation}
\label{sup:eq:main_ineq}
    |Y_{\text{linear}} - Y_\alpha|^2 \leq (1-\alpha)^2(\sigma(W_2W_1)^2|X|^2+|W_2b_1|^2)
\end{equation}
Using $x^\delta$ and $\delta$ as defined in equation  \eqref{sup:eq:xdelta} we have:

\begin{equation}
\label{sup:eq:prob1}
\begin{aligned}
     P\{ &(1-\alpha)^2(\sigma(W_2W_1)^2|X|^2+|W_2b_1|^2) \leq \\
     &(1-\alpha)^2(\sigma(W_2W_1)^2|x^\delta|^2+|W_2b_1|^2)\} > 1-\delta
\end{aligned}
\end{equation}

Using equations \eqref{sup:eq:prob1} and \eqref{sup:eq:main_ineq} with using the $C$ defined in \eqref{sup:eq:xdelta} we have:
\begin{equation}
\label{sup:eq:p_error_final}
    P\left\{ |Y_{\text{linear}} - Y_\alpha|^2 \leq C\times(1-\alpha)^2 \right\} > 1 - \delta
\end{equation}

\section{AdamLC}
\label{sup:adam}
Inspired by AdamW \cite{loshchilov2017decoupled} we decouple the adam optimizer with our regularization so the adaptive learning rate does not make the regularization un-predictable the algorithm is given in \ref{alg:adamlc}. In this algorithm $lc$ is the regularization constant.

\begin{algorithm}[H]
\caption{AdamLC Optimization Algorithm}
\label{alg:adamlc}
\begin{algorithmic}[1]
\REQUIRE $\alpha, \beta_1, \beta_2, \epsilon, \theta_0, lc$
\ENSURE Optimized parameters $\theta$
\STATE Initialize time step $t \gets 0$
\STATE Initialize first moment vector $m_0 \gets \mathbf{0}$
\STATE Initialize second moment vector $v_0 \gets \mathbf{0}$
\STATE Initialize parameter vector $\theta \gets \theta_0$
\WHILE{not converged}
    \STATE $t \gets t + 1$
    \STATE $g_t \gets \text{gradient at } \theta$
    \STATE $m_t \gets \beta_1 \cdot m_{t-1} + (1 - \beta_1) \cdot g_t$
    \STATE $v_t \gets \beta_2 \cdot v_{t-1} + (1 - \beta_2) \cdot g_t^2$
    \STATE $\hat{m}_t \gets \frac{m_t}{1 - \beta_1^t}$
    \STATE $\hat{v}_t \gets \frac{v_t}{1 - \beta_2^t}$
    \STATE $\theta \gets \theta - \alpha \cdot \frac{\hat{m}_t}{\sqrt{\hat{v}_t} + \epsilon}$
    \IF{$\theta$ has a negative slope}
        \STATE $\theta \gets \theta - 2 \cdot (\theta - 1) \cdot lc \cdot \alpha$
    \ENDIF
\ENDWHILE
\STATE \textbf{return} $\theta$
\end{algorithmic}
\end{algorithm}

\section{Experiment Details}
\label{sup:experiments}
The experimental files are accessible at [https://github.com/sohail2810/LayerCollapse].

\subsection{Regularization Performance of \MethodName}

Training was conducted using a modified version of the Adam optimizer, as detailed in Appendix \ref{sup:adam}. A consistent regularization parameter, Regularization 5, was applied across all experiments, and the same value was used for comparisons with the \textit{LayerDrop} method. The learning rate scheduler LiveTune \cite{shabgahi2023livetune} was employed, with an initial learning rate set to $0.001$.

In all experiments, a batch size of 128 was maintained, and the models were trained for a total of 3,000 steps. For performance evaluation, we applied a stride of 32 on the complete test dataset of Wikitext-103.

\subsection{\MethodName for Compression}

\subsubsection{Performance on GLUE}
Because of the diversity of tasks each have their own requirements for training (i.e. optimizer, learning rate, and epochs)
All tasks except for QQP the SGD optimizer is used, and the modified adamLC \ref{sup:adam} is used for QQP.
In all experiments initial learning rate is 0.0001 which reduces to 0.00001 linearly.
Collapsing is done one layer at a time. Starting from the final un-collapsed layer. In each step we reduce the regularization loss of the active layer during a retraining phase, when the loss is bellow $1e^{-4}$ we move to the next layer. 
All tasks for Bert-base use batch size 128. Bert-large is trained on batch size 20.

\subsubsection{Performance of ImageNet-1k}
\label{sup:sub:standalone_compression}
\noindent\textbf{Dataset:} 
We utilized the full ImageNet-1k dataset, with image dimensions of $224\times224$. Training images were augmented with random cropping, horizontal flipping, and rotation (up to 15 degrees). Validation images were not augmented. Normalization factors were model-specific: standard and mean values were set to $0.5$ for all color channels in most models, except for the MLP-Mixer, which used the default ImageNet normalization (mean = ($0.482, 0.456, 0.406$), std = ($0.229, 0.224, 0.225$)). The batch size was fixed at 256 for all experiments.

\noindent\textbf{Self Distillation:}
We accelerated finetuning by using the original model’s predicted labels, akin to Knowledge Distillation techniques. The loss function combined cross-entropy and KL Divergence terms:

\begin{equation}
\label{sup:eq:loss}
\begin{aligned}
    loss = \frac{1}{2}CrossEntropy(p, \hat{p}) + \frac{1}{2}&KLDivergance(p, \Tilde{p})\\
    & + Regularization
\end{aligned}
\end{equation}

where $\hat{p}$ and $\Tilde{p}$ represent true and original model’s predicted labels, respectively. Regularization is as defined in Eq. \eqref{eq:regularizer}.

Layer collapsing was performed sequentially, starting from the last layer. Each layer’s MLP was replaced with our CollapsibleMLP, featuring PReLU activations, and finetuned using Eq. \eqref{sup:eq:loss} for up to 10 epochs. Following this, we applied the collapse procedure (Algorithm \ref{alg:collapse}). A regularization strength of $0.05$ and an SGD optimizer (momentum $0.9$, initial learning rate $0.0005$) were used. After 5 epochs, the learning rate was reduced to $10^{-4}$.

\subsection{LayerCollapse vs. Knowledge Distillation}
\label{sup:sub:lc_vs_kd}
For both LayerCollapse and Knowledge Distillation, the target architectures were identical. The LayerCollapse setup followed the procedure outlined in Section \ref{sup:sub:standalone_compression}. 
The distillation process employed the SGD optimizer with an initial learning rate of $0.001$, reduced to $0.0001$ upon plateau, and concluded when no further progress was possible.

\subsection{Study of Regularization on VGG}
\label{sup:sub:ablation}
To ensure comparability, all models, including those without LayerCollapse, shared the same training settings. The SGD optimizer with a starting learning rate of $0.005$ was used, decreasing by a factor of $0.3$ at epochs 75, 150, and 250. A dropout probability of $0.5$ was applied. In LayerCollapse experiments, $0.4$ fraction of layers were regularized with a strength of $0.05$. Hyper-parameter tuning was not a focus of this study.



\end{document}